%% file: main.tex
\documentclass[10pt,twocolumn,letterpaper]{article}

\usepackage[review]{cvpr}      
\usepackage{makecell}
\usepackage{caption}
\usepackage{float}

\input{preamble}

\renewcommand{\thefootnote}{\fnsymbol{footnote}}

\definecolor{cvprblue}{rgb}{0.21,0.49,0.74}
\usepackage[pagebackref,breaklinks,colorlinks,citecolor=cvprblue]{hyperref}

\def\paperID{7763} 
\def\confName{CVPR}

\title{DiffInDScene: Diffusion-based High-Quality 3D Indoor Scene Generation}

\author{{Xiaoliang Ju}$^{1}$\footnotemark[1], {Zhaoyang Huang}$^{1}$\footnotemark[1], {Yijin Li}$^{2}$, {Guofeng Zhang}$^{2}$, {Yu Qiao}$^{3}$, {Hongsheng Li}$^{1}$\\
$^1$ MMLab, The Chinese University of Hong Kong   $\quad$  $^2$ Zhejiang University $\quad $$^3$ Shanghai AI Laboratory\\
{\tt\small \{akira,drinkingcoder\}@link.cuhk.edu.hk,  hsli@ee.cuhk.edu.hk}
}

\begin{document}
\begin{figure}
\twocolumn[{%
\renewcommand\twocolumn[1][]{#1}
\maketitle
\begin{center}
    \centering
    \captionsetup{type=figure}
    \includegraphics[trim={1.9cm 5cm 2cm 5cm},clip,width=\textwidth]{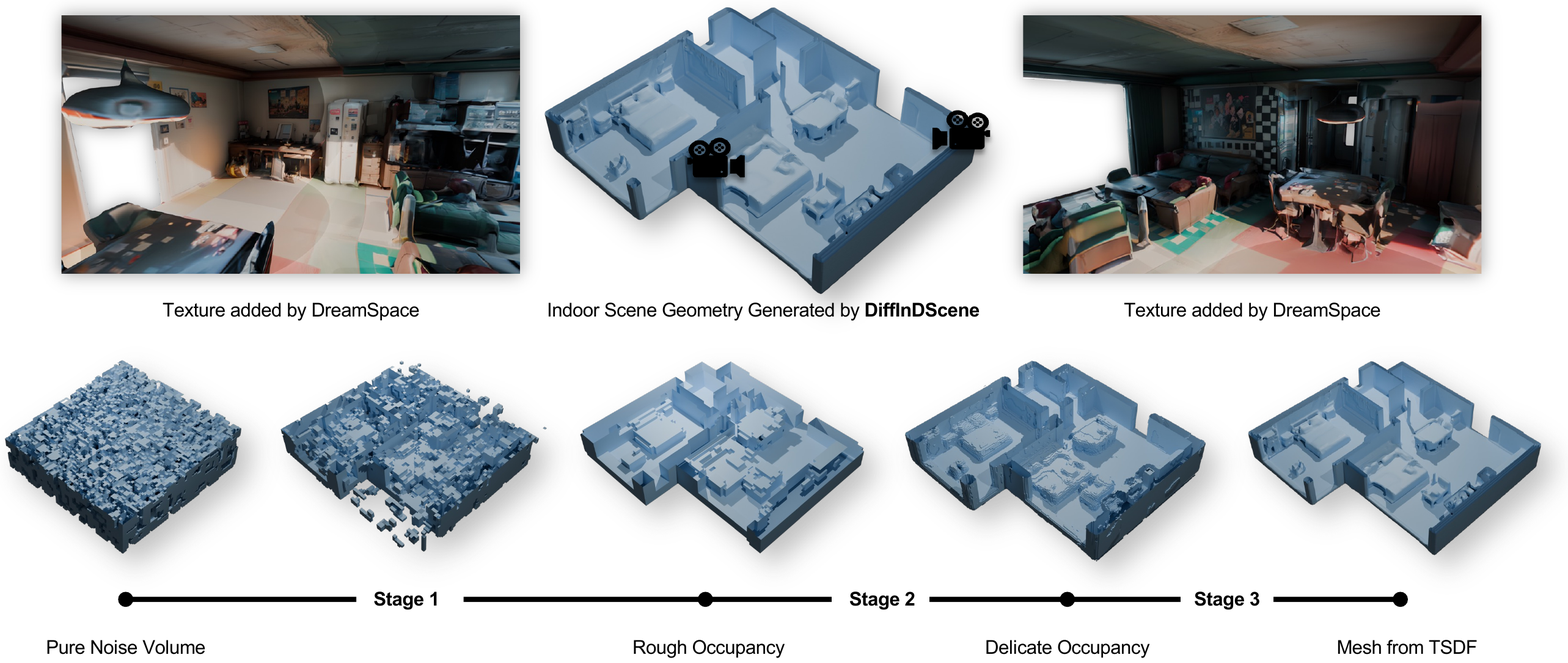}
    \caption{Coarse-to-fine indoor scene geometry generation using a sparse diffusion framework. For better visualization, the texture is produced by {\it DreamSpace}~\cite{yang2023dreamspace} after the scene geometry is generated by our DiffInDScene.}
    \label{fig:teaser}
\end{center}%
}]
\end{figure}

{\let\thefootnote\relax\footnotetext{{$^*$ Joint first authorship}}}

\input{sec/0_abstract}    
\input{sec/1_intro}
\input{sec/2_related_works}

\input{sec/3_method}

\input{sec/4_experiment}

\input{sec/5_conclusion}

{
    \small
    \bibliographystyle{ieeenat_fullname}
    \bibliography{main}
}
\input{sec/X_suppl}

\end{document}

%% file: preamble.tex
\usepackage[dvipsnames]{xcolor}


%% file: sec/0_abstract.tex
\begin{abstract}
We present DiffInDScene, a novel framework for tackling the problem of high-quality 3D indoor scene generation, which is challenging due to the complexity and diversity of the indoor scene geometry.
Although diffusion-based generative models have previously demonstrated impressive performance in image generation and object-level 3D generation, they have not yet been applied to room-level 3D generation due to their computationally intensive costs.
In DiffInDScene, we propose a cascaded 3D diffusion pipeline that is efficient and possesses strong generative performance for Truncated Signed Distance Function (TSDF).  The whole pipeline is designed to run on a sparse occupancy space in a coarse-to-fine fashion.
Inspired by KinectFusion's incremental alignment and fusion of local TSDF volumes, we propose a diffusion-based SDF fusion approach that iteratively diffuses and fuses local TSDF volumes, facilitating the generation of an entire room environment. The generated results demonstrate that our work is capable to achieve high-quality room generation directly in three-dimensional space, starting from scratch.
In addition to the scene generation, the final part of DiffInDScene can be used as a post-processing module to refine the 3D reconstruction results from multi-view stereo.
According to the user study, the mesh quality generated by our DiffInDScene can even outperform the ground truth mesh provided by ScanNet. Please visit our project page for the latest progress and
demonstrations: \textcolor{magenta}{\url{https://akirahero.github.io/diffindscene/}}.
\end{abstract}

%% file: sec/1_intro.tex
\section{Introduction}
\label{sec:intro}

3D scene production is a fundamental task in 3D computer vision with many applications, such as Augmented Reality (AR), game development and embodied AI~\cite{puig2023habitat}, where the quality of 3D scene geometry plays a paramount role. While the 3D reconstruction from multi-view stereo~\cite{neuralrecon,transformerfusion} can recover scenes from real-world, the quality of the resultant meshes is far from satisfactory, for the major mesh details might be lost during iterative fusion.
Recently, diffusion models~\cite{denoising_diffusion,improved_diffusion} have shown their great ability in generating images and objects of high quality. 
Here we want to ask a question: \textit{Can we exploit diffusion models to produce 3D scenes}?
In this paper, 
we propose a novel framework DiffInDScene, which not only helps optimize the results of 3D reconstruction, but also generates high-quality indoor spatial geometry from scratch (see Fig.~\ref{fig:teaser}).

Diffusion models are a class of generative models designed for synthesizing data by iterative denoising.
The popular Denoising Diffusion Probabilistic Model (DDPM) training paradigm starts the denoising from pure Gaussian noise.
Training diffusion models for room-level Truncated Signed Distance Function (TSDF) volumes is challenging because of its large size. Previous 3D diffusion models only focus on object-level 3D generations.
As reported by InstantNGP~\cite{instantngp}, only 2.57\% voxels are informative in common 3D scenes.

To deal with the large scale of indoor scenes, we propose a coarse-to-fine sparse diffusion pipeline consists of multiple stages. The first few stages are used to generate the occupancy volume in a coarse-to-fine manner, and the last one generates the TSDF values in the sparsely occupied voxels.  For the first few occupancy generation stages, a multi-scale auto-encoder is designed to encode occupancy to latent space, providing feature guidance for occupancy generation. This approach allows us to employ latent diffusion and further compress the size of input volumes. 
Additionally, we propose a sparse 3D diffusion model denoising only on the sparsely occupied voxels of TSDF or the occupancy latent volumes, which saves two orders of computational and memory costs.

Although the cascaded sparse diffusion pipeline significantly reduces the required computational resources, 
the considerable variation in room sizes still poses a challenge when attempting to directly train on a room-level TSDF volume, at the final stage of our cascaded diffusion process. To increase the data variation within each mini-batch, we randomly crop local TSDF volumes of smaller sizes from the original large volume for training. During inference, we design a stochastic TSDF fusion algorithm that generates the entire room by iterative denoising and fusing local TSDF volumes. The fusion method allows for the generation of a complete and unified TSDF within a large scene while efficiently decomposing the scene into smaller crops, thereby conserving computing resources.
Our proposed diffusion method can be also utilized to refine indoor scene meshes such as the reconstruction results from multi-view stereo, such as NeuralRecon~\cite{neuralrecon}. Given the TSDF volume of reconstructed scene as occupancy condition, our diffusion model can effectively refine and optimize the TSDF volume towards the ground truth.

Our contributions can be summarized as four-fold: 1) We propose a novel framework DiffInDScene for room-level indoor scene generation with a sparse diffusion model that saves two orders of resource consumption. 2) We design a multi-scale auto-encoder to provide feature guidance for the scene occupancy generation. 3) We propose a novel algorithm that fuses diffusion-based local TSDF volumes, which enables large-scale indoor scene generation. 4) DiffInDScene exhibits a promising capability in producing high-quality room-level geometry through both generation from scratch and refinement of existing reconstructions.

%% file: sec/2_related_works.tex
\section{Related Works}
\label{sec:related_works}

\noindent \textbf{Diffusion Models.}
The diffusion model~\cite{nonequilibrium,denoising_diffusion,improved_diffusion} has emerged as a promising class of generative models for learning data distributions through an iterative denoising process.
They have shown impressive visual quality in diverse applications of 2D image synthesis, encompassing image inpainting~\cite{diffusion_repaint}, super-resolution~\cite{diffusion_super_reso,diffusion_cascaded}, editing~\cite{diffusion_sdedit}, text-to-image synthesis~\cite{diffusion_text2image_1,diffusion_text2image_2}, and video generation~\cite{diffusion_video_1,diffusion_video_2}.
Nevertheless, the application of diffusion models in the 3D domain has received limited attention in comparison to the extensive exploration seen in the 2D domain. In the 3D domain, existing research has focused on the generation of individual objects~\cite{diffusion_obj_1,diffusion_obj_2,diffusion_obj_3,diffusion_obj_4}, while less attention has been paid to the synthesis of entire scenes, which possess significantly higher levels of semantic and geometric complexity, as well as the expansive spatial extent present in 3D scene synthesis.

\noindent \textbf{3D Shape Generation.} Extensive exploration has been conducted on various 3D representation methods, such as voxel~\cite{wu2016learning, xie2018learning}, point cloud~\cite{wu2023sketch,nichol2022point,yang2019pointflow}, and implicit field~\cite{shim2023diffusion, li2023diffusion}, in conjunction with different generation models such as diffusion models~\cite{shim2023diffusion} and GANs~\cite{wu2016learning}.  While these approaches have shown success in object-level generation, transferring them to scene generation at a larger scale is still challenging. Firstly, the large scale leads to exponential growth in computation resources consumption in training and inference processes. Additionally, object-level generation is comparatively easier due to simpler geometries and less diversity.

\noindent \textbf{3D Scene Synthesis.} In recent decades, the field of 3D scene synthesis has experienced extensive investigation, particularly driven by the proliferation of 3D indoor scene datasets~\cite{matterport3d,scannet} and advancements in 3D deep learning~\cite{pointnet,pointnet++,point_voxel_net}.
However, current methods mainly focus on synthesizing plausible 3D scene arrangements~\cite{obj_arr_1,obj_arr_2,obj_arr_3,obj_arr_4}. They usually learn to synthesize the scene graph as the intermediate scene representation and retrieve objects from available dataset. 
In contrast to these methods, we aim to simultaneously synthesize both the scene arrangement and the detailed geometry. 
Text2Room~\cite{hollein2023text2room} is the most related work to ours in recent years, which leverages pre-trained 2D text-to-image models to synthesize a sequence of images and then conduct an iterative reconstruction. While such method can produce room-scale geometry, the results are often fragmentary and distorted, limiting their practical applications in areas such as gaming or AR.

%% file: sec/3_method.tex
\begin{figure*}[!t]
    \centering
    \includegraphics[trim={1.7cm 6.5cm 1.7cm 3.5cm},clip,width=1.0\textwidth]{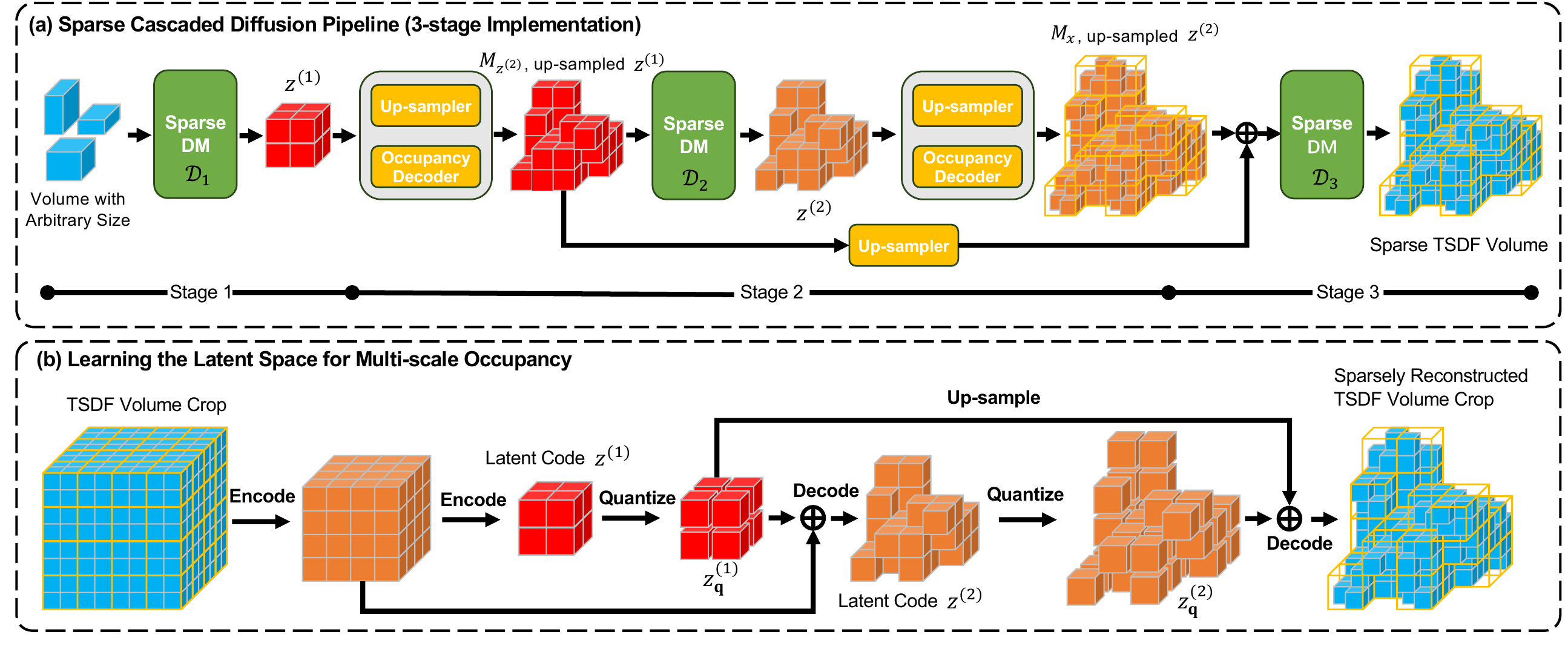}
    \caption{Sparse cascaded diffusion with a multi-scale occupancy embedding.}
    \label{fig:pipeline}
\end{figure*}
\section{Methodology}

\subsection{Overview}
To generate room-level 3D geometry, the greatest challenge is the large scale, as it requires substantial computing resources.
We employ a cascaded diffusion model to generate the whole room in a coarse-to-fine manner. The first stage is to generate the coarse structure of the whole room. The following stages further refine the rough shape to a 3D occupancy field with higher resolutions. At the final stage, the resolution increases to the highest level, and we crop the whole scene to overlapped pieces to generate the final detailed Truncated Signed Distance Function (TSDF) volume. In every stage, we use a separate sparse diffusion model to reduce the resource consumption, which exclusively denoises on sparsely distributed occupancy. In our implementation, we use 3 stages to create indoor geometry up to size of $512\times 512\times 128$.

Such cascaded solution has three advantages. First, the computation resource consumption is constrained in all stages. Second, compared with piece-wise generation or incremental generation methods, the first stage of our model is able to sketch the global structure of the scene, which helps to generate a complete layout with unified and detailed structure. Third, every stage can be trained independently, and the generation process can be stopped in advance when generating an unsatisfied layout at early stage.

\subsection{Cascaded Diffusion for Indoor Geometry Generation}
\label{subsec:cascaded}

We propose a sparse cascaded diffusion model as shown in Fig.~\ref{fig:pipeline} (a). 
In our implementation, a 3-stage diffusion process is utilized to generate a complete indoor scene starting from noise.
The first 2 stages are utilized to generate and refine the 3D binary scene occupancy volume, and the final is used to generate TSDF value within the occupancy. As TSDF volume only retains information near the object surface, we simply define all voxels containing valid TSDF values as the binary scene occupancy.

Assume we have multi-scale occupancy embeddings $z^{(1)}$, $z^{(2)}$ of a TSDF volume $x$ with increasing resolutions, and their binary occupancy masks are $M_{z^{(1)}}$, $M_{z^{(2)}}$, $M_{x}$, satisfying
\begin{equation}
\label{eq:m1}
        M_{z^{(2)}}= G_1\left(z^{(1)}, M_{z^{(1)}}\right),
\end{equation}

\begin{equation}
\label{eq:m2}
        M_{x} = G_2\left(z^{(1)}, z^{(2)},M_{z^{(2)}}\right),
\end{equation}
where $G_1, G_2$ are occupancy decoders, and we will explain them in detail in Section~\ref{sec:ms_occ_encoder} together with the occupancy latents $z^{(1)}$, $z^{(2)}$. Then the 3-stage diffusion processes $\{\mathcal{D}_1,\mathcal{D}_2,\mathcal{D}_3\}$ can be established as follows.

\begin{itemize}
    \item Stage 1 generates the occupancy latent code  $z^{(1)}$ of the lowest resolution. Given a fixed volume  $z^{(1)}_T$ and its occupancy mask $M_{z^{(1)}_T}$, the diffusion process performs denoising operation in $T$ timesteps to obtain $z^{(1)}_0$ as 
    \begin{equation}
        z^{(1)}_0 = \mathcal{D}_1(z^{(1)}_T, M_{z^{(1)}_T}),
    \end{equation}
    where $z^{(1)}_T$ is filled with Gaussian noise. By defining mask $M_{z^{(1)}_T}$ according to datasamples, we can include data samples with varying volume sizes in each mini-batch without padding or cropping, and control over the maximum area for scene generation.

    \item Stage 2 generates latent code of higher resolution $z^{(2)}$ conditioned on $z^{(1)}$ as
    \begin{equation}
        z^{(2)}_0 = \mathcal{D}_2(z^{(2)}_T, M_{z^{(2)}_T}; z^{(1)}_0),
    \end{equation}
    where $M_{z^{(2)}_T}$ is obtained utilizing Eq.~\eqref{eq:m1}, and $z^{(2)}_T$ is filled with another Gaussian noise volume.
    \item Stage 3 generates the final TSDF volume with all those generated latent codes $z^{(1)}, z^{(2)}$ as input conditions
    \begin{equation}
        x_0 = \mathcal{D}_3(x_T, M_{x_T}; z^{(1)}_0, z^{(2)}_0),
    \end{equation}
    where $M_{x_T}$ is obtained by Eq.~\eqref{eq:m2}, and $M_{x_T}$ is filled with Gaussian noise.
\end{itemize}
Compared with generating occupancy directly, the occupancy embedding can guide the refinement of the occupancy throughout the diffusion process, particularly at the initial stage of generation. 

\noindent \textbf{Sparse Diffusion.} We follow DDPM~\cite{nonequilibrium,denoising_diffusion} to implement the sparse diffusion.
Each stage of our model employs a separate sparse diffusion, with the only distinction lies in the types of input and output. In this section, we use $v$ to represent any kind of volumes such as TSDF $x$ and latent codes $z^{(1)}, z^{(2)}$,  and $y$ to denote the diffusion condition with $M$ as their shared sparsity mask.

DDPM~\cite{nonequilibrium,denoising_diffusion} transforms a sample volume $v_0$, to a white Gaussian noise $v_T\sim\mathcal{N}(0,1)$ in $T$ steps. 
In each step $t$, 
the sample $v_t$ is obtained by adding i.i.d. Gaussian noise with variance $\beta_t$ and scaling the sample in the previous step $v_{t-1}$ with $\sqrt{1-\beta_t}$:
\begin{equation}
q\left(v_t \mid v_{t-1}\right)=\mathcal{N}\left(v_t ; \sqrt{1-\beta_t} v_{t-1}, \beta_t \mathbf{I}\right),
\end{equation}
which is also called the forward direction. 
On the other hand, the reverse process can be depicted as:
\begin{equation}
p_\theta\left(v_{t-1} \mid v_t, y, M\right)=\mathcal{N}\left(v_{t-1} ; \mu_\theta, \Sigma_\theta\right),
\end{equation}
where $y$ denotes the extra condition.
Diffusion models are trained to reverse the forward process, predicting   $\mu_\theta$ as

\begin{equation}
\mu_\theta\left(v_t, y, M, t\right)=\frac{1}{\sqrt{\alpha_t}}\left(v_t-\frac{\beta_t}{\sqrt{1-\bar{\alpha}_t}} \epsilon_\theta\left(v_t, y, M, t\right)\right),
\end{equation}
where $\alpha_t:=1-\beta_t$, $ \bar{\alpha}_t:=\prod_{s=0}^t \alpha_s$, and $\epsilon_\theta$ is the neural network with the parameter set $\theta$, which has a UNet-like structure. With the occupancy mask $M$,  $\epsilon_\theta$ denoises only sparsely occupied voxels with sparse convolutions and attentions. We implement $\epsilon_\theta$ using the engine of TorchSparse~\cite{tang2022torchsparse}. The details of network architecture are provided in the supplementary materials.

A mean square error loss masked by $M$ is used to supervise the noise prediction as 
\begin{equation} 
L_{\rm diff}=E_{t, v_0, \epsilon, y, M} \left( M \odot \left[\left\|\epsilon-\epsilon_\theta\left(v_t, y, M, t\right)\right\|_2^2\right]\right).
\end{equation}

\subsection{Learning the Latent Space for Multi-scale Occupancy}
\label{sec:ms_occ_encoder}
To obtain hierarchical occupancy embeddings and their decoders, we design a multi-scale Patch-VQGAN as Fig.~\ref{fig:pipeline}(b), inspired by VQ-VAE-2~\cite{razavi2019generating}. 
The encoder takes TSDF volume $x$ as input, and outputs the occupancy embedding $z^{(1)}, z^{(2)}$. These embeddings are expected to be decoded to the occupancy mask $M_{z^{(2)}}$ and $M_{x}$ as Eq.~\eqref{eq:m1} and Eq.~\eqref{eq:m2}, where ground truth of $M_{z^{(2)}}$ is obtained by downsampling from $M_{x}$ via maxpooling.
To capture more shape details, we also add a TSDF decoding head.
After training, the encoder is no longer utilized, and only the occupancy decoders are employed to convert the latent code to occupancy.

Formally, we define the encoder as $E$, the decoders as $G$, and the element-wise quantization of latent volume as $\mathbf{q}(\cdot)$. Any ground truth TSDF volume $x\in\mathbb{R}^{H\times W \times L}$ can be encoded progressively into
$(z^{(1)},z^{(2)}) = E(x)$, where $z^{(1)},z^{(2)}\subseteq \mathbb{R}^{d}$. 
The quantization is formulated as
\begin{equation}
    \mathbf{q}(z) := (\arg\min_{z_p\in Z} \|z_{ijk} - z_p \|) \in \mathbb{R}^{h\times w \times l \times d},
\end{equation}
where $Z=\{z_k\}_{k=1}^K \subset \mathbb{R}^{d}$ denotes the discrete codebook of size $K$, and $z_{ijk}$ represents the latent vector at coordinate $i,j,k$ of the latent volume. 
For simplicity, we define $z^{(1)}_\mathbf{q}:=\mathbf{q}(z^{(1)})$, $z^{(2)}_\mathbf{q}:=\mathbf{q}(z^{(2)})$ in Fig.~\ref{fig:pipeline}(b).

During the training process, we randomly crop cubes of $96\times96\times96$ from the original TSDF volumes as data samples, so that diverse crops from different scenes are contained in each mini-batch.
The training loss is defined in Eq.~\eqref{eq:loss}, where $L_{\rm rec}$, $L_{\rm vq}$, $L_{\rm GAN}$ denote the reconstruction loss, vector quantization loss and the adversarial loss from a simple multi-layer discriminator, 
\begin{equation}
    \label{eq:loss}
    L= L_{\rm rec} + \lambda_1 L_{\rm vq} + \lambda_2 L_{\rm GAN},
\end{equation}
where $\lambda_1$ and $\lambda_2$ are hyper parameters to weight losses of different types.
For the encoder-decoder training, we use a $L_1$ loss to supervise the TSDF value, and binary cross-entropy (BCE) to supervise the occupancy masks, as shown in Eq.~\eqref{eq:loss-rec}. 
\begin{equation}
\label{eq:loss-rec}
        L_{\rm rec}={\rm BCE}(\hat{M_x}, M_x)+{\rm BCE}(\hat{M_{z^{(2)}}}, M_{z^{(2)}})+\|x-\hat{x}\|_1
\end{equation}
The $L_{\rm vq}$, $L_{\rm GAN}$ are defined similar to ~\cite{esser2021taming}.

\subsection{Local Fusion for Global Diffusion}
\label{subsec:local_fusion}

The volume cropping operation in the training of the final stage diffusion raises a question: how to infer on a complete scene using a model trained on crops? The independent generation in a crop-by-crop manner as image generators such as ~\cite{diffusion_repaint,diffusion_text2image_2} may cause inconsistent results between adjacent crops. 
To address this issue, a fusion algorithm is proposed to perform the joint diffusion process concurrently on the overlapping local volumes.

During inference, we split the indoor space into $K$ overlapping 3D crops $\{\mathcal{P}^0,\mathcal{P}^1,\dots,\mathcal{P}^{K-1},\}$ that cover the entire room, and generate the TSDF for the entire room by concurrently diffusing the $K$ crops with stochastic fusion.
We denote the global TSDF at the timestep $t$ as $x_t$ and the $k$-th crop at the timestep $t$ as $x^k_t$ and the global TSDF at the timestep $t$ as $x_t$. 
At time step $t$, we need to obtain the global TSDF $x_t$ by fusing local TSDFs $x^k_t$ and then update local TSDFs from the global TSDF: $x^k_t(\mathbf{p}_i)=x_t(\mathbf{p}_i)$.
After synchronizing local TSDFs with fusion, each crop step to the next time step individually.
Specifically, for a voxel grid $\mathbf{p}$, suppose $\mathcal{G}(\mathbf{p})$ contains the crops that cover $\mathbf{p}$: $\mathcal{G}(\mathbf{p}):=\{k|\mathbf{p}\in \mathcal{P}^k\}$, we need to obtain $x_t(\mathbf{p})$ by fusing the crops $\{x^k_t(\mathbf{p})|k\in \mathcal{G}(\mathbf{p})\}$ overlapping on $\mathbf{p}$.

\begin{figure}[!t]
    \centering
    \includegraphics[trim={3.0cm 13cm 8.5cm 3.5cm},clip,width=.9\textwidth]{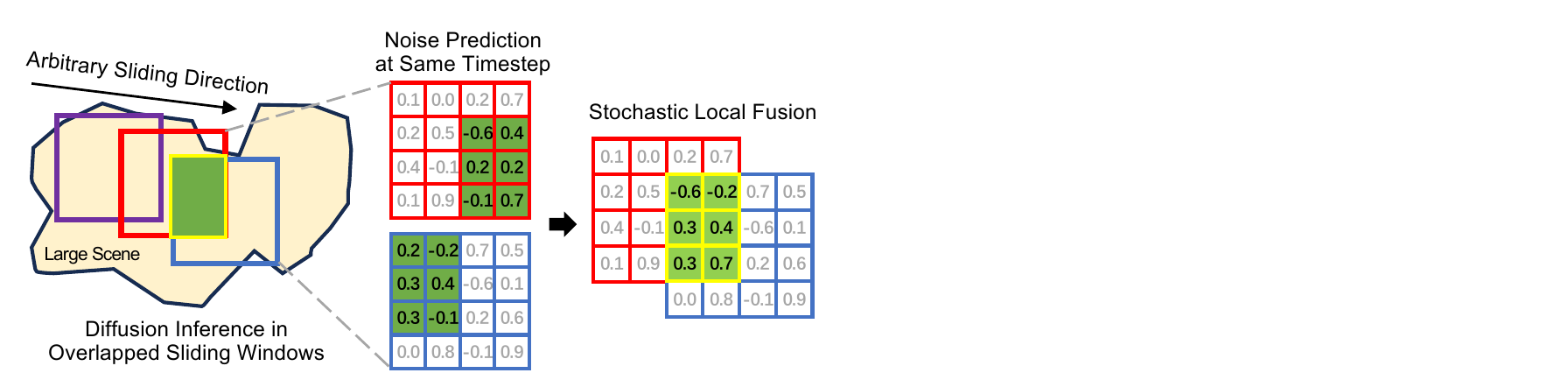}
    \caption{Stochastic TSDF Fusion.}
    \label{fig:local_fusion}
\end{figure}
\noindent \textbf{Stochastic TSDF Fusion.} A straightforward fusion algorithm is taking the average TSDFs of the local crops:  $x_t(\mathbf{p})=\frac{1}{|\mathcal{G}(\mathbf{p})|}\sum_{k\in \mathcal{G}} x_t^k(\mathbf{p})$, which is also adopted by the classical KinectFusion~\cite{kinectfusion}. However, average fusion significantly reduces the variance of the sample distribution. To ensure the generation quality and global consistency, we propose stochastic fusion to keep the distribution and fuse TSDFs in the reverse process. An example of 2-windows fusion is shown in Fig.~\ref{fig:local_fusion}.
Specifically, we randomly sample an index $k$ from $\mathcal{G}(\mathbf{p})$ in a uniform distribution to update the global TSDF
$x_t(\mathbf{p})=x^k_t(\mathbf{p})$, which remains the distribution:
\begin{equation}
    \begin{aligned}
        x_t(\mathbf{p}) \sim \mathcal{N}(\mu_t^k(\mathbf{p}), \Sigma^k_t(\mathbf{p})), k={\rm RandomSelect}\left( \mathcal{G}(\mathbf{p}) \right ).
    \end{aligned}
\end{equation}

\subsection{Extension to Refining 3D Reconstruction}

Given a rough geometry represented by the occupancy volume of a scene, the TSDF volume can be generated
by directly adopting the final stage of our cascaded diffusion framework.

\begin{figure}[!t]
    \centering
    \includegraphics[trim={2.2cm 12.5cm 12.8cm 4cm},clip,width=.5\textwidth]{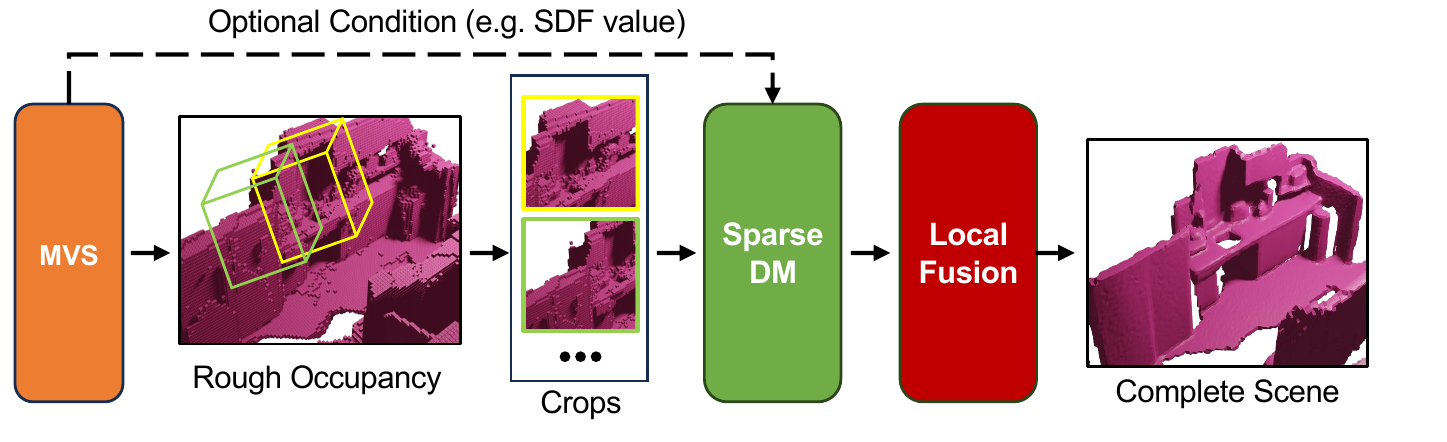}
    \caption{Refine or recover scene from MVS methods.}
    \label{fig:recon}
\end{figure}

The most direct application is to recover/refine scenes produced by multi-view stereo methods or even LiDAR mappers as depicted in Fig.~\ref{fig:recon}, as obtaining an occupancy of the scene is relatively straightforward using Multi-view Stereo (MVS) techniques. If the MVS method provides a TSDF result, it can be utilized as the diffusion condition to our cascaded diffusion network. The diffusion process is applied to the crops of the occupancy volume as inputs, and the results are fused using our local fusion module at each timestep, so that we can obtain a complete and refined scene reconstruction. 

%% file: sec/4_experiment.tex
\section{Experiment}

DiffInDScene is a versatile tool capable of generating detailed indoor scene geometry at the room level. It is not only capable of creating scenes from scratch but also has the ability to refine or recover scenes using occupancy fields produced by multi-view stereo methods. To train DiffInDScene for scene generation, we utilized the 3D-FRONT dataset~\cite{fu20213d} provided by Alibaba, which consists of 6813 furnished houses with meshes. For training purposes, we selected 5913 houses with dimensions smaller than $512\times 512\times 128$, with a voxel size of 0.04m.
For scene refinement or recovery on multi-view stereo (MVS), we trained DiffInDScene using the training split of the ScanNet dataset, with the NeuralRecon~\cite{neuralrecon} as the MVS module. The ScanNet dataset contains 1613 indoor scenes, each accompanied by ground-truth camera poses and surface reconstructions. To ensure consistency, we adopted the same data split as NeuralRecon for training and evaluation purposes.

\subsection{Indoor 3D Scene Generation}

In this section, we will present the results of our indoor scene generation, both quantitatively and qualitatively. To the best of our knowledge, Text2Room stands as the state-of-the-art solution to generate 3D geometry of room-level indoor scenes. Since there are few other works that can directly generate scene structures and output the corresponding mesh models, here we 
also compare with "Text2Room + Poisson" to enrich the experimental comparison, which means a Poisson reconstruction~\cite{kazhdan2006poisson} is added as a refinement of Text2Room.
\label{sec-exp: generation}

\noindent \textbf{Metrics.} We employ mesh quality as a measure to reflect the overall quality of the generated 3D scene geometry.
Regarding mesh quality, we noticed that noisy meshes and high-quality meshes exhibit distinct distributions of triangle shapes. Noisy meshes and problematic regions are characterized by triangles with low aspect ratio, circularity, and shape regularity, as introduced in~\cite{BRANDTS20082227,cignoni2008meshlab}. Therefore, these 3 factors are selected as the objective metric of mesh quality.

In addition, we perform a user study similar to the one conducted in Text2Room~\cite{hollein2023text2room} as the subjective metric on the mesh quality, including the completeness and perceptual quality. 
The scores range from 1 to 5, with higher scores indicating a better alignment with the evaluation metrics.

\noindent \textbf{Quantitative Results.} For the objective metrics, We randomly select 11 scenes from different methods, and calculate the average value of all triangular mesh faces. For the user study, we randomly choose 5 scenes generated by each method and gather ratings from a group of 40 users with basic knowledge in 3D modeling.
We summarize the evaluation results as Table~\ref{tab: gen_mesh_quality} shows.
Our method demonstrates superior performance on those metrics compared to the the other approaches. While employing Poisson reconstruction as a refinement step enhances the results of Text2Room, it does not lead to a significant improvement.
\begin{figure*}[!ht]
    \centering
    \includegraphics[trim={2.8cm 3.5cm 9cm 3.5cm},clip,width=1\textwidth]{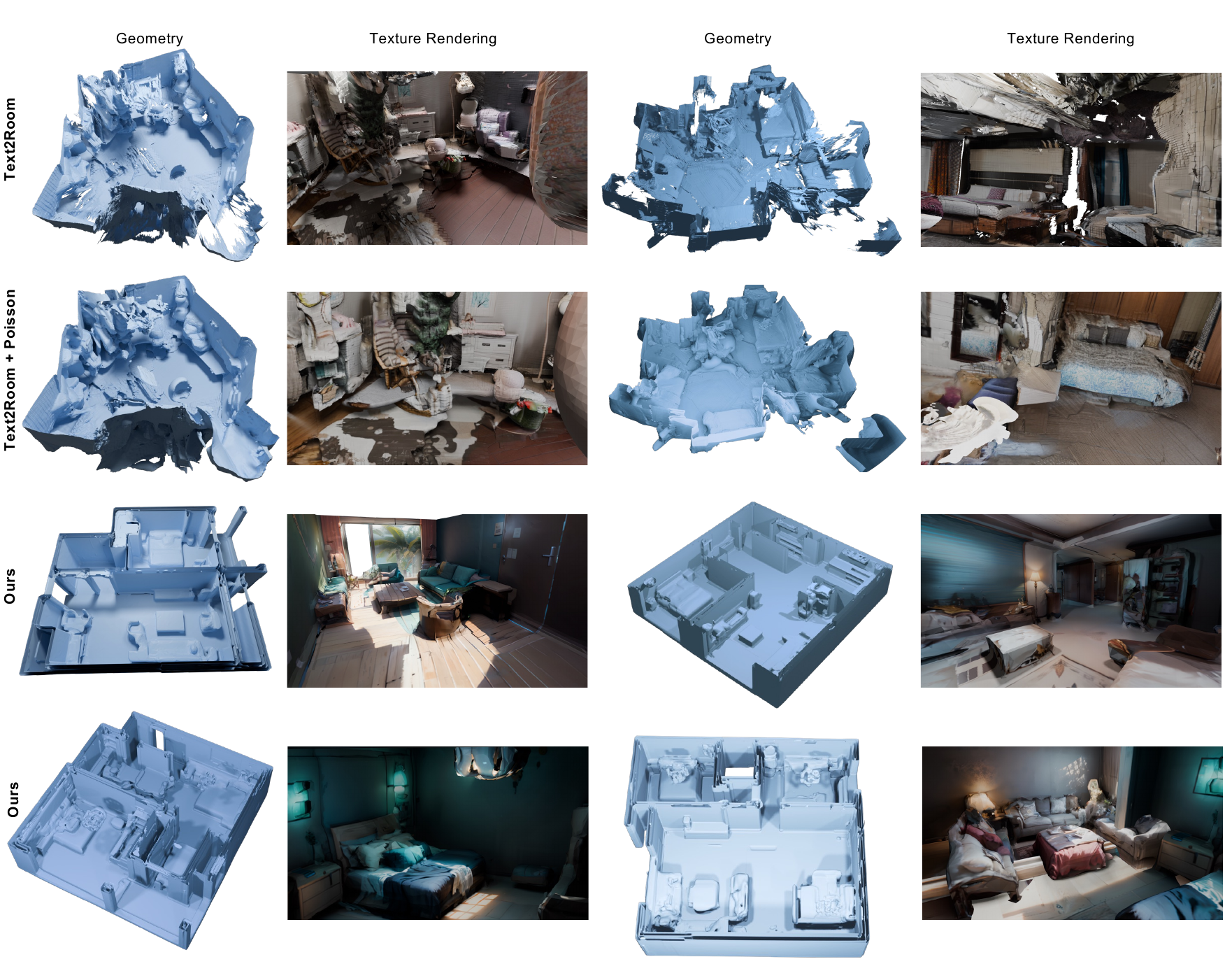}
    \caption{Comparison of indoor scene generation. The texture renderings of our results are produced by DreamSpace~\cite{yang2023dreamspace}. The ceilings are removed from the original meshes for a better visualization.}
    \label{fig:generated_scenes}
\end{figure*}
\begin{figure}[!ht]
    \centering
    \includegraphics[trim={3.5cm 7cm 8.6cm 7cm},clip,width=\columnwidth]{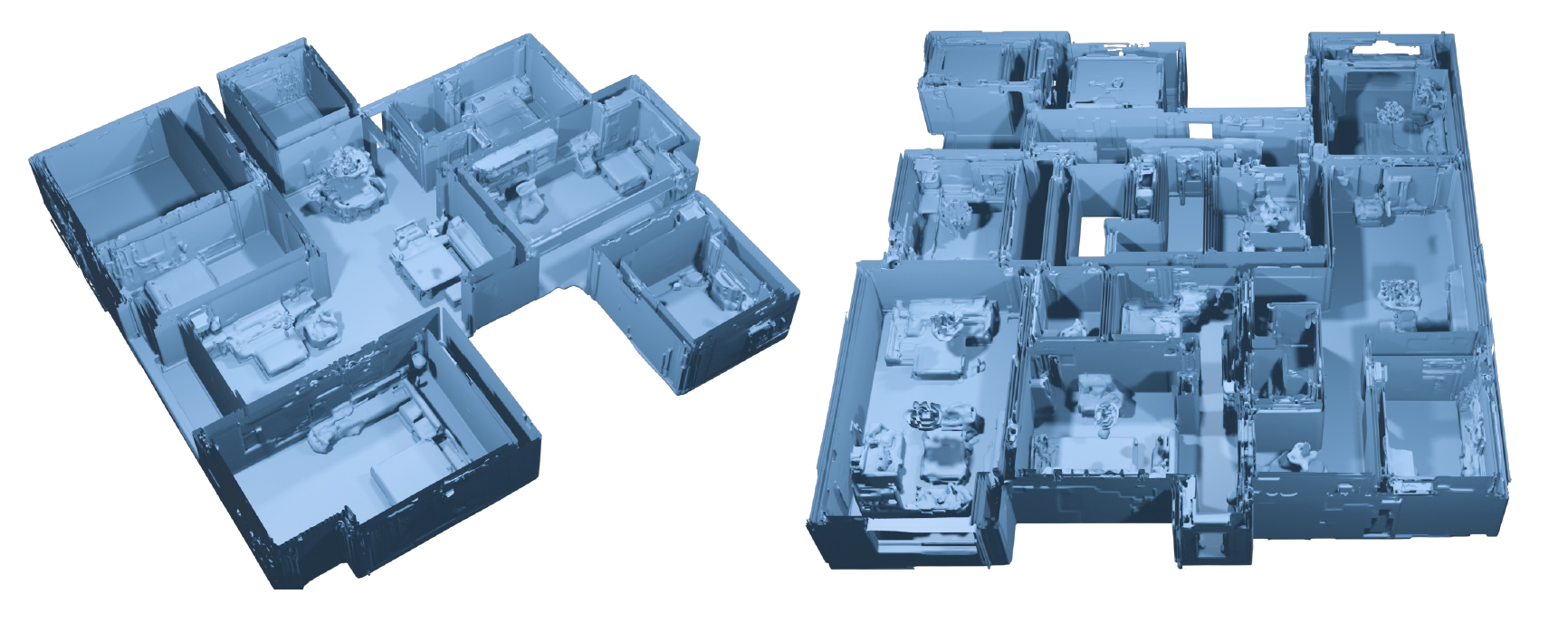}
    \caption{Samples with more complex structures generated by DiffInDScene. Note that the floating objects in this figure are ceiling lamps, which are reserved after we remove the ceilings for visualization. }
    \label{fig:larger_generated_scenes}
\end{figure}
\begin{table}[bt]
    \centering
    \caption{
    Geometry quality comparison, including mesh quality (Aspect Ratio, Circularity, and Shape Regularity) and user study on the completeness and perceptual quality. }
    \resizebox{\columnwidth}{!}{
    \begin{tabular}{lcccccc}
\toprule
             & \makecell{Text2Room~\cite{hollein2023text2room}}& \makecell{Text2Room\\+ Poisson~\cite{kazhdan2006poisson}} & Ours\\ \hline

Aspe. mean$\uparrow$  &0.416 & 0.443 &	\textbf{0.473}  \\
Aspe. var$\downarrow$ &0.022 &  0.029 & \textbf{0.009}  \\
Circ. mean$\uparrow$ &0.674 & 0.709 &	\textbf{0.781}  \\
Circ. var$\downarrow$ &0.052 &0.057&	\textbf{0.022} \\
Shap. mean$\uparrow$ &0.716 &  0.730	 & \textbf{0.816}\\
Shap. var$\downarrow$ &0.045 & 0.060	 & \textbf{0.023} \\
\midrule
Completeness &2.532& 3.228&\textbf{4.856}\\
Perceptual &2.472& 2.812& \textbf{4.836}\\
\bottomrule
\end{tabular}

}
\label{tab: gen_mesh_quality}
\end{table}

\noindent \textbf{Qualitative Results.} The generated scene samples from different methods are listed in Fig.~\ref{fig:generated_scenes}. We compare both un-textured meshes and textured scene renderings in our evaluation. While Text2Room is capable of reconstructing the scene from images, it often results in serious distortions and fragmentation. As indicated by our quantitative evaluation, applying Poisson reconstruction helps in filling the holes, but it has limited impact on improving the overall structure of the scene. Our method can produce larger rooms with clearer structures and more complex layout. Furthermore, by incorporating DreamSpace as a post-processing step, we can achieve high-quality scene renderings. This combination allows for enhanced visual output and improved overall scene representation.

Compared with the middle-size house showed in Fig.\ref{fig:generated_scenes}, two larger and more complex samples are shown in Fig.~\ref{fig:larger_generated_scenes}. Additional samples showcasing larger views can be found in the supplementary materials.

\subsection{Ablation Study}

\begin{figure}[!t]
    \centering
    \includegraphics[trim={3.5cm 4.5cm 12cm 3.2cm},clip,width=\columnwidth]{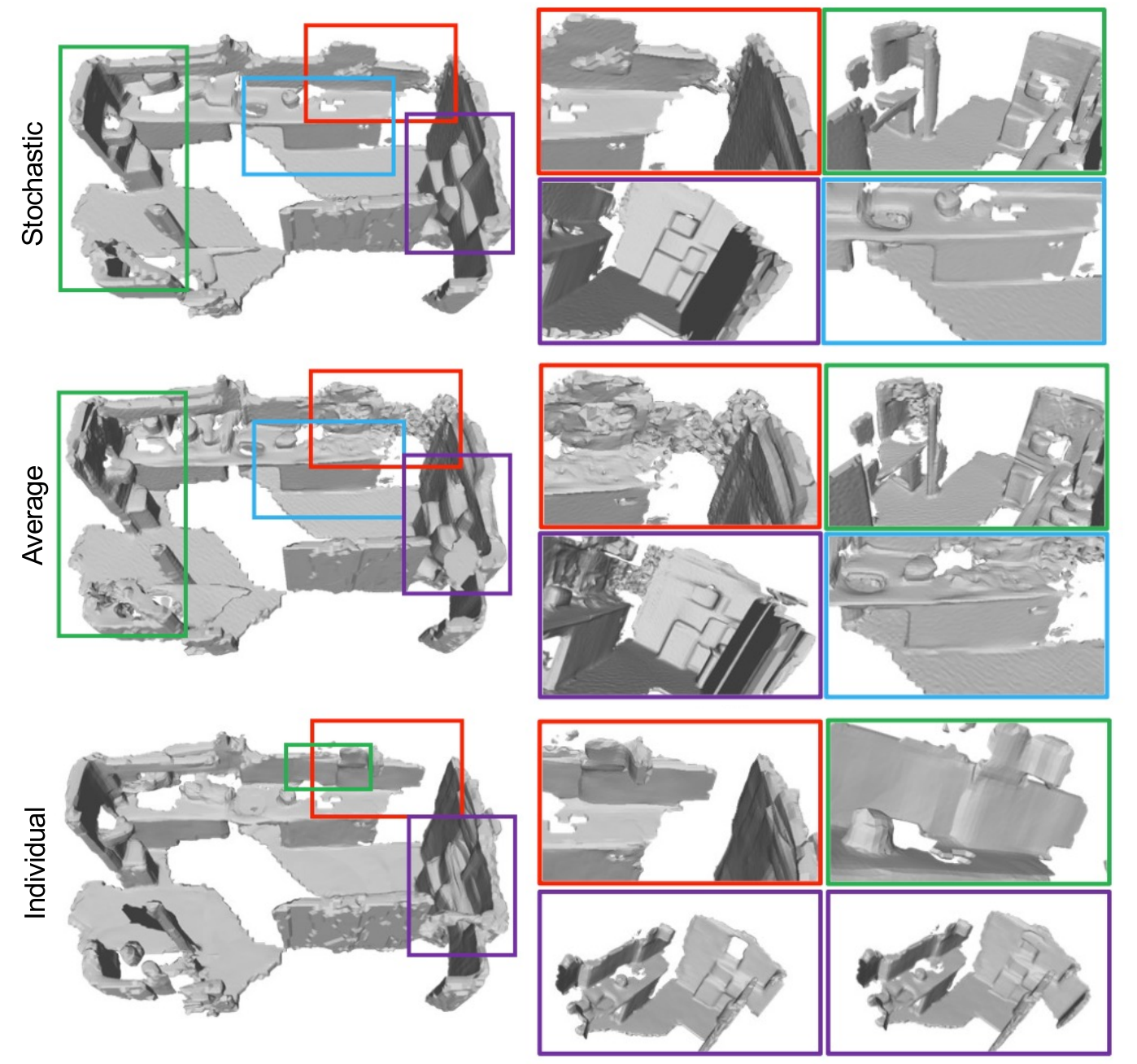}
    \caption{Comparison of different fusion methods in the final stage of our cascaded diffusion.}
    \label{fig:fusion_ablation}
\end{figure}
\noindent \textbf{Sparse or Dense.}
To compare the dense diffusion and the sparse diffusion, we implement two networks using sparse and dense convolution respectively, with exactly the same structures. 
Several randomly cropped TSDF volumes $(96\times 96 \times 96)$ from ScanNet dataset are fed into these two models. The resource consumption of the two strategies are shown in Table~\ref{tab:ablation_sparse}.
This experiment is conducted on the platform equipped with RTX3090 GPU with 24GB memory, with the diffusion model running in training status.
The sparse diffusion requires fewer resources and runs faster with fewer parameters, due to the characteristic of the occupancy distribution. In our test data, the largest occupancy rate of the TSDF crops are less than 20\%. 
\begin{table}[]
\centering
\caption{Resource Consumption Comparison.}
\resizebox{1.0\columnwidth}{!}{
\begin{tabular}{lccccc}
\toprule
           & TFLOPs  & Parameters(M) & \multicolumn{3}{l}{GPU Memory(GB)} \\
Batch Size & 1      & 1          & 1        & 2        & 4        \\ \hline
Sparse     & 0.008 & 161.5     & 11.8    & 15.3    & 22.8    \\
Dense      & 3.290  & 161.5     & 22.8    & -        & -        \\
\bottomrule
\end{tabular}
}
\label{tab:ablation_sparse}
\end{table}

\noindent \textbf{Diffusion with Fusion.}
To compare the different fusion methods mentioned in Section~\ref{subsec:local_fusion}, we perform the final stage of our cascaded diffusion model on the given occupancy from a scene of ScanNet dataset. The results are shown in Fig.~\ref{fig:fusion_ablation}. 
Individual diffusion presents significant inconsistency between adjacent crops. Average fusion generates meshes of lower quality because the sample distribution during diffusion is disturbed. Our stochastic diffusion remains global consistency and generates high-quality meshes.

\subsection{Refinement and Recovery on MVS}
\label{sec-exp:refine}

\begin{figure}[!ht]
    \centering
    \includegraphics[trim={14.5cm 3cm 4.3cm 3cm},clip,width=\columnwidth]{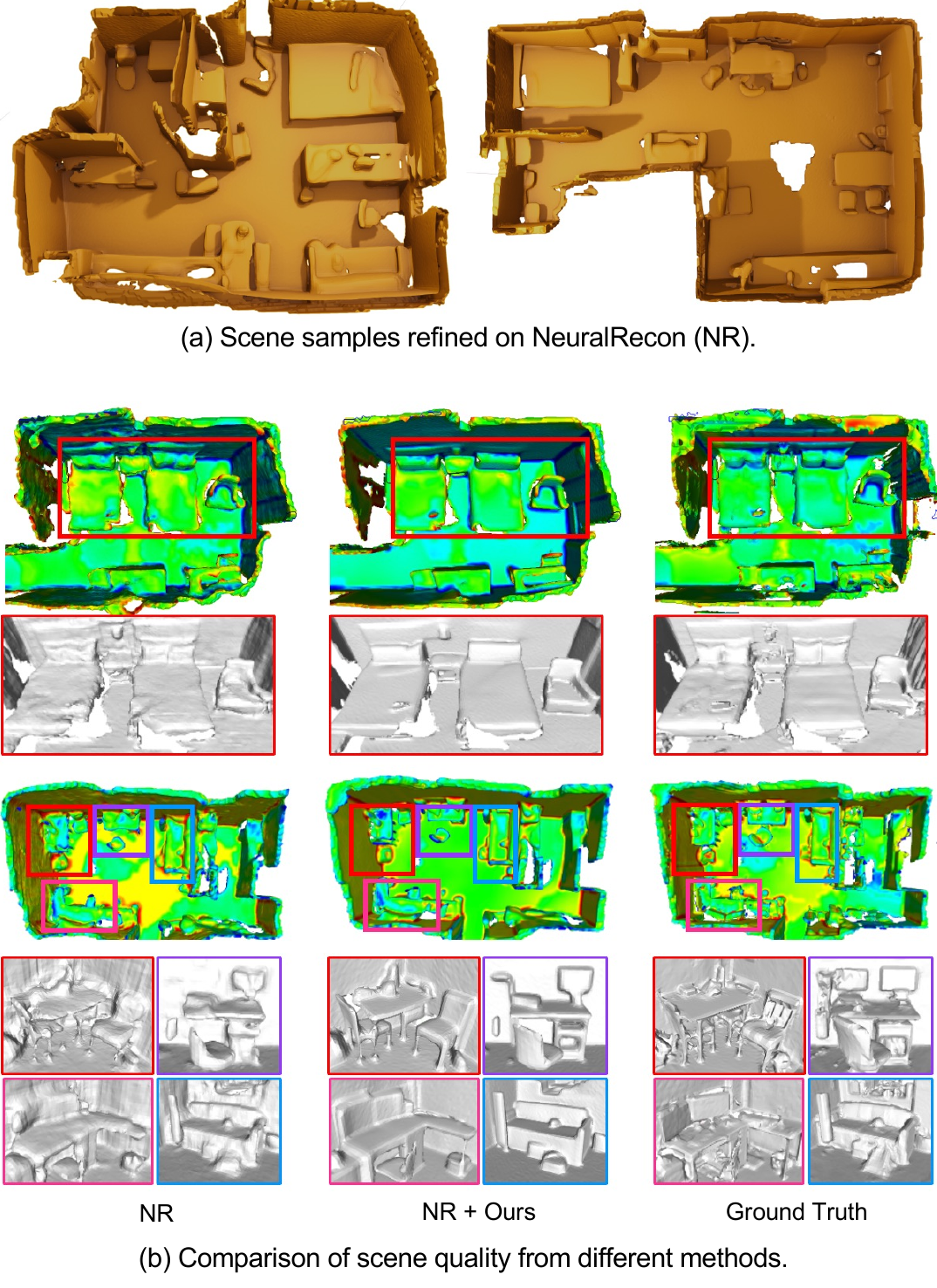}
    \caption{Sample scene reconstructions on ScanNet dataset. The meshes are colored according to curvatures in sub-figure(b), where green regions denote lower curvatures.}
    \label{fig:rec_mesh_samples}
\end{figure}
\noindent \textbf{Metrics.} Our objective is to obtain high-quality 3D scenes from MVS, aiming for results comparable to the ground truth in various aspects, including details, completeness, tightness, and sharpness.
As a generative model, the proposed method operates within a relaxed occupancy, which may cause drift from the ground truth. As a result, conventional correspondence-based metrics for 3D reconstruction are not suitable. Instead, we employ metrics unaffected by the relaxed occupancy, such as normal error distribution and mesh quality. 

Normal error is used to evaluate the similarity of surface orientation between the reconstructed and ground truth meshes. Smaller normal errors indicate better alignment to the ground truth. We filter out outliers with errors larger than $90^\circ$ and analyze the percentage of inlier normals below a threshold ($<T^\circ$ ratio) and the mean normal error of those inliers.  We use the same mesh quality metrics as Section~\ref{sec-exp: generation}.

\noindent \textbf{Results.} The samples of scene reconstruction are shown in Fig.~\ref{fig:rec_mesh_samples}.
 As Table~\ref{tab:normal error} shows, our method significantly outperforms the other three methods on all thresholds of normal error. 
 As for the mesh quality, we compute the mean and variance of the three scores described above and compare them in Table~\ref{tab: mesh quality}.
Our results presents significantly better performance than the other three methods, i.e., higher scores with smaller score variance.
Moreover, we also include ground truth mesh for comparison. Interestingly, our results also outperforms the ground truth except for the aspect ratio score, which indicates the high quality of our reconstructed meshes.

\begin{table}[!t]
\centering
\caption{Normal error comparison. ``NR'', ``Lap'', and ``SR'' are the abbreviations of NeuralRecon, Laplacian and SimpleRecon, respectively.}
\resizebox{\columnwidth}{!}{
\begin{tabular}{lcccc}
\toprule
             & NR~\cite{neuralrecon} & \begin{tabular}[c]{@{}l@{}}NR + Lap~\cite{laplacian_edit}\end{tabular} & SR~\cite{simplerecon} & NR + Ours \\ \midrule

$<90^\circ$mean$\downarrow$   & 34.65 &	37.12&	35.44&	\textbf{30.4}    \\
$<90^\circ$ ratio$\uparrow$   & 100\%  & 100\%                                                                    & 100\%                                                                  & \textbf{100\%}   \\
$<45^\circ$ mean$\downarrow$   & 10.27&	12.34	&12.51	&\textbf{8.09} \\
$<45^\circ$ ratio$\uparrow$   & 59.97\%	&57.97\%&	60.20\%&	\textbf{65.05\%} \\
$<30^\circ$ mean$\downarrow$   &6.45&	8.17&	8.31&	\textbf{5.05}   \\
$<30^\circ$ ratio$\uparrow$  & 52.88\%&	49.89\%	&51.67\%&	\textbf{59.27\%}  \\
\bottomrule
\end{tabular}
}
\label{tab:normal error}
\end{table}

\begin{table}[t]
    \centering
    \caption{Mesh quality comparison. We compare the mean and variance of three scores: Aspect Ratio, Circularity, and Shape Regularity. NR(occ) means only the occupancy is used, without the conditional TSDF depicted by the dash line in Fig.~\ref{fig:recon}.}
    \resizebox{\columnwidth}{!}{
    \begin{tabular}{lcccccc}
\toprule
             & NR~\cite{neuralrecon} & \makecell{NR +\\ Lap~\cite{laplacian_edit}} & SR~\cite{simplerecon} & \makecell{NR(occ)\\+ Ours} & \makecell{NR +\\Ours} & GT\\ \hline

Aspe. mean$\uparrow$  & 0.459&	0.437&	0.436&	0.469 & 0.457 &	\textbf{0.477}  \\
Aspe. var$\downarrow$ & 0.022 &	0.023&	0.024& 0.020 & \textbf{0.016}&	0.022  \\
Circ. mean$\uparrow$ & 0.740 &	0.712 &	0.708 & 0.742 &	\textbf{0.763} &	0.758 \\
Circ. var$\downarrow$ &0.041&	0.052&	0.054& 0.041 &	\textbf{0.030}&	0.034  \\
Shap. mean$\uparrow$ &  0.772	&0.746	&0.739&	0.793 & \textbf{0.797}&	0.793\\
Shap. var$\downarrow$ & 0.041&	0.052&	0.055&	0.041 & \textbf{0.030}&	0.031 \\
\bottomrule
\end{tabular}

}
\label{tab: mesh quality}
\end{table}

\noindent \textbf{User Study.}
The user study is conducted to rank scene quality, providing subjective evaluations to complement the objective metrics. We randomly select 10 scenes from the test split of the ScanNet dataset and generate reconstructed meshes using four methods. Users are asked to rank the methods based on details, completeness, plane quality, and edge quality. We collect ranking results from 51 users.
As shown in Table~\ref{tab:user_study}, our reconstructed mesh significantly outperforms NeuralRecon, ``NeuralRecon+Laplacian Denoising" (denoted as "NR + Lap"), and even surpasses the quality of the ground truth meshes.

\begin{table}[!t]
\centering
\caption{User study on the refinement of scene reconstruction.}
\resizebox{\columnwidth}{!}{

\begin{tabular}{lcccc}
\toprule
             & NR~\cite{neuralrecon} & \begin{tabular}[c]{@{}l@{}}NR + Lap~\cite{laplacian_edit}\end{tabular} & NR + Ours & GT \\ \hline
Details$\uparrow$      & 12.26       & 6.80                                                                     & 17.41                                                                  & \textbf{25.50}        \\
Completeness$\uparrow$ & 11.15       & 8.84                                                                     & \textbf{21.30}                                                                  & 20.76        \\
Tight Plane$\uparrow$  & 5.48        & 12.10                                                                    & \textbf{25.85}                                                                 & 18.17        \\
Sharp Edge$\uparrow$   & 8.22        & 10.20                                                                    & \textbf{22.58}                                                                  & 21.19   \\
Overall (Sum)$\uparrow$   & 37.10        & 37.98                                                                    & \textbf{87.14}                                                                  & 85.62   \\
\bottomrule
\end{tabular}
}
\label{tab:user_study}
\end{table}

%% file: sec/5_conclusion.tex
\section{Conclusion}
We have presented DiffInDScene as a novel framework for diffusion-based high-quality indoor scene generation. DiffInDScene mainly consists of three modules: 1) a sparse diffusion network that efficiently denoises 3D volumes on occupied voxels, 2) a multi-scale Patch-VQGAN for occupancy decoding, 3) a cascaded diffusion framework to generate room-level scene from scratch, and 4) a stochastic fusion algorithm for diffusion-based local TSDFs, which enables large-scale indoor scene generation.
In the future, we will explore to generate scenes with various conditions such as text and sketch. 

%% file: sec/X_suppl.tex
\clearpage
\setcounter{page}{1}
\maketitlesupplementary

\section{Video Demonstration}
To gain a more comprehensive understanding of our method for generating the indoor scene, we kindly invite you to watch the attached video. The video demonstrates an example of the coarse-to-fine generation process, and the the post-processing of texturing using DreamSpace~\cite{yang2023dreamspace}. Furthermore, to provide a more detailed and complete perspective on the inner scene structures, a random walk is conducted within the generated scene.

\section{Implementation Details}
\subsection{Dataset and Preprocessing}
\noindent \textbf{Indoor Scene Generation from Scratch.}
3D-FRONT~\cite{fu20213d} provides professionally designed layouts and a large number of rooms populated by high-quality 3D models. However, when organizing the mesh models to a complete scene, the meshes may intersect with each other. Additionally, most of them are not watertight meshes. These factors lead to erroneous Truncated Signed Distance Function (TSDF) volumes. In such cases, the meshes retrieved from TSDF volumes contains lots of wrong connections. To address this problem, we perform a solidification and voxel remeshing on each scene mesh, using a pipeline of modifiers from Blender with a voxel size of 0.02m. All meshes are saved as triangular format.
After the watertight meshes are obtained, we derive the SDF volumes by using a open-source software SDFGen~\cite{batty2015sdfgen}, with a resolution of 0.04m. Then the SDF volumes are truncated to TSDF by a maximum distance of 0.12m.

\noindent \textbf{Refinement on the Reconstruction from Multi-view Stereo(MVS).}
We use the official train / validation / test split of ScanNet(v2) dataset, including 1201 / 312 / 100 scenes respectively. For there is no TSDF ground truth provided in this dataset, we adopt a TSDF fusion method like~\cite{kinectfusion} to produce the ground truth as NeuralRecon does. We only use TSDF data without any other data type such as images in the whole training/testing process. To compare the reconstruction results with pretrained NeuralRecon, the grid size of TSDF volume is set to 0.04m, and the truncation distance is set to 0.12m. The default value of the TSDF volume is 1.0.

In the training process, a random volume crop of $96\times96\times 96$ is used as data augmentation, where a random rotation between $\left[0, 2\pi\right]$ and a random translation is performed before cropping. To ensure that the sampling crop contains sufficient occupied voxels, the translation is limited in the bounding box of global occupied region, and the entire cropped volume should be within the boundary of this region.

\subsection{Sparse Diffusion Model}
\begin{figure*}[!t]
    \centering
    \includegraphics[trim={3.7cm 3cm 3.7cm 2.5cm},clip,width=\textwidth]{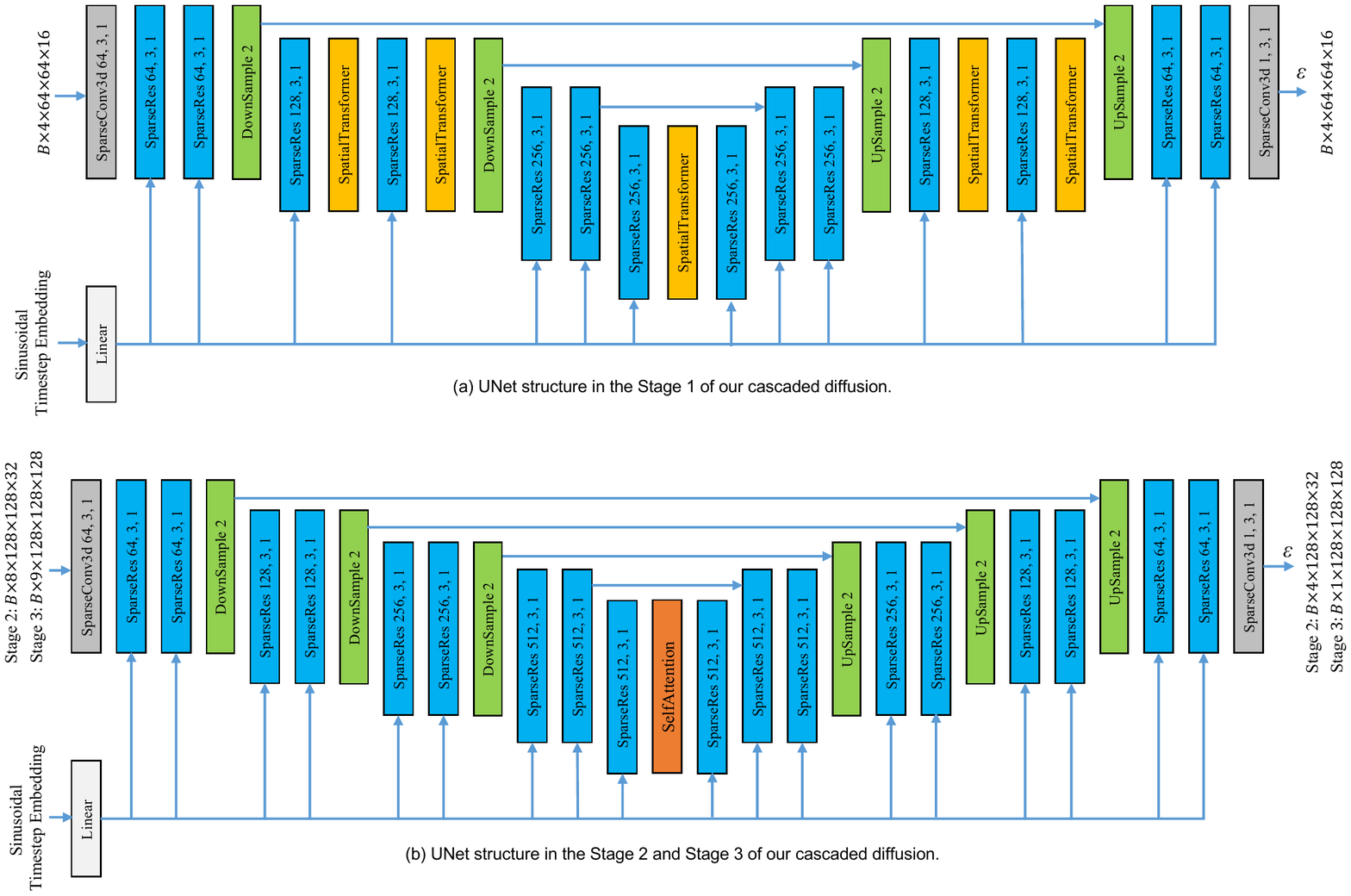}
    \caption{Noise prediction networks in our cascaded diffusion. In Stage 1, we use multiple Spatial
    Transformers as (a) shows. 
    In Stage 2 and Stage 3, we use same network structure as (b), with only one attention layer in the middle of network.}
    \label{supfig:diffusion_net}
\end{figure*}
\begin{figure}
    \centering
    \includegraphics[trim={7.cm 9cm 9cm 7cm},clip,width=0.5\textwidth]{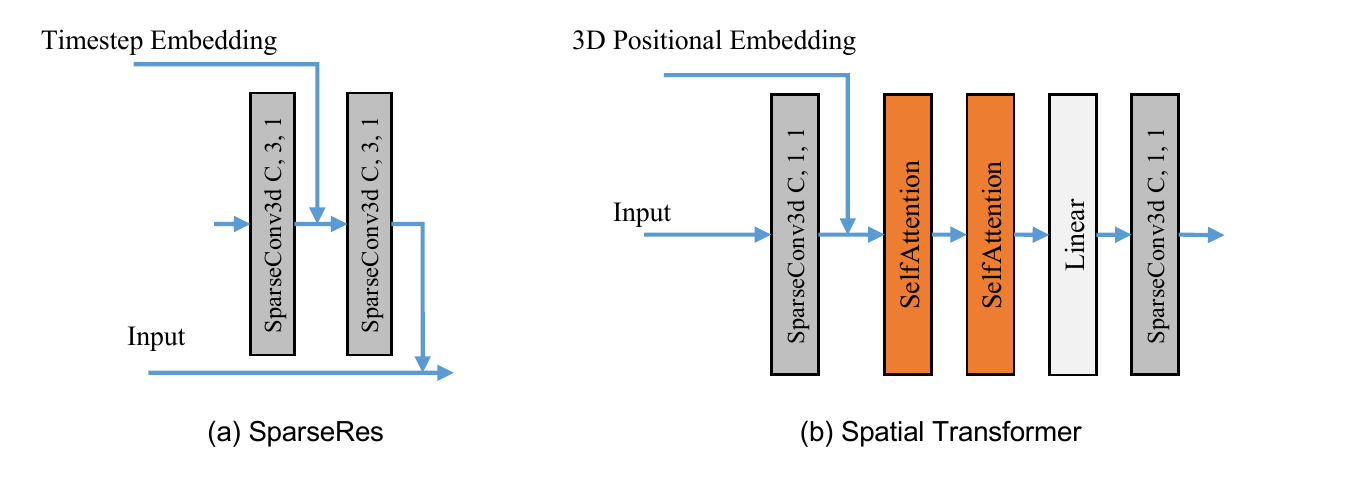}
    \caption{Sparse units widely used in our implementation of noise prediction network in sparse diffusion.}
    \label{supfig:sparse_units}
\end{figure}
\noindent \textbf{Network Structure.}
\texttt{TorchSparse}~\cite{tang2022torchsparse} is used to implement the UNet structure of our network for noise prediction. A group normalization(32 groups) and a SiLU activation are used successively before any layer of sparse convolution. 
The network strctures used in difference stages of our cascacded diffusion are shown in Fig.~\ref{supfig:diffusion_net}, where SparseRes and Spatial Transformer are key components of our implementation as shown in Fig.~\ref{supfig:sparse_units}.

\noindent \textbf{Training \& Inference Settings.}
The network parameters are randomly initialized in training process, and we use the Adam optimizer with a learning rate of $1.0 \times 10^{-4}$. 

As for the diffusion framework, the \texttt{DDIMScheduler} in the open-source diffusers~\cite{von-platen-etal-2022-diffusers} is developed as our code-base.
Following~\cite{chen2020wavegrad} and~\cite{saharia2022palette}, we adopt the $\alpha-$conditioning to stabilize training, and enable the parameter tuning over the noise schedule and the timesteps during inference stage. More concretely, the cumulative product of $\alpha_t$ namely $\bar{\alpha}_t$ is used as a substitute of the timestep $t$ as time embedding in most existing works. 
In Section~\ref{sec-exp: generation}, we use a cosine noise schedule with $2000$ timesteps during training, and the same noise schedule is used with $200$ time-steps during inference within the DDIM framework.
In Section~\ref{sec-exp:refine}, we use a linear noise schedule of $(1e-6, 0.01)$ with $2000$ timesteps during training, and the same noise schedule is used with $100$ time-steps during inference within the DDIM framework. 
The clip range for TSDF sampling is $\left[-3.0, 3.0\right]$.

\subsection{PatchVQGAN for Learning the Occupancy Embedding}
\begin{figure*}[!t]
    \centering
    \includegraphics[trim={3cm 3.5cm 5cm 3cm},clip,width=\textwidth]{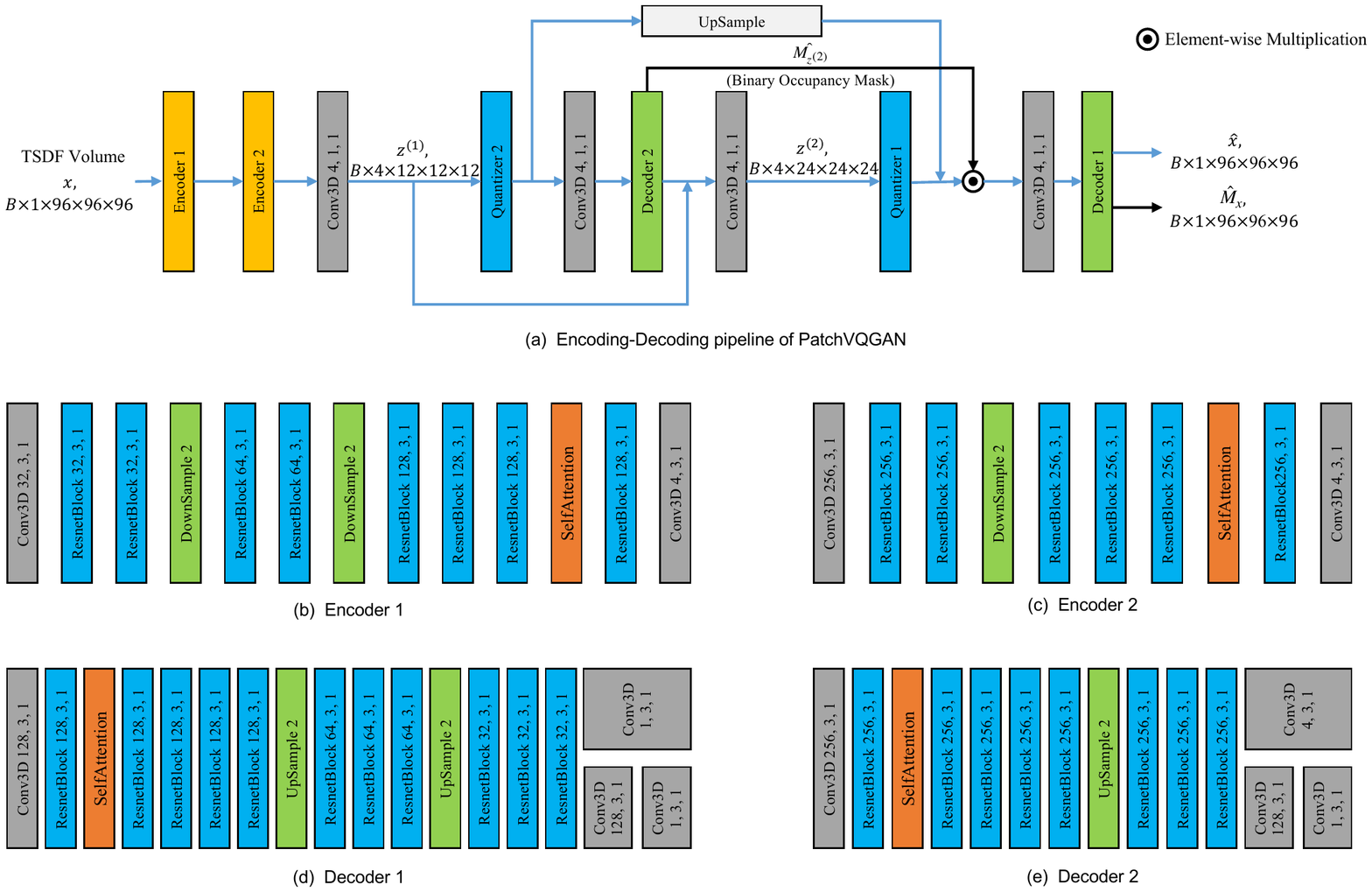}
    \caption{Network structure of PatchVQGAN.}
    \label{supfig:vqgan}
\end{figure*}

\noindent \textbf{Network Structure.} The network structure of PatchVQGAN described in Section~\ref{sec:ms_occ_encoder} is shown in Fig.~\ref{supfig:vqgan}. The multi-scale encoding and decoding processes are slightly coupled with each other, while we simplify the description of the whole model for better understanding in Section~\ref{sec:ms_occ_encoder}. The encoder and decoder are implemented hierarchically as "Encoder 1", "Encoder 2", "Decoder 1", and "Decoder 2" as shown in Fig.~\ref{supfig:vqgan} (b)-(e). The multi-layer feed-forward discriminator is omitted here.

Different from~\cite{esser2021taming}, we use quantizers with Gumbel-Softmax~\cite{jang2016categorical} which enables a differentiable discrete sampling. The size of codebook is 8192, with the embedding dimension of 4 as commonly adopted in ~\cite{esser2021taming}\cite{rombach2022high}.

\noindent \textbf{Training \& Inference Settings.}
The hyper parameters in Eq.~\eqref{eq:loss} are initially set to $\lambda_1=1.0$, $\lambda_2=0.2$. Additionally, a dynamic weight adapting strategy as ~\cite{esser2021taming} is employed to control $\lambda_2$.
The network parameters are randomly initialized with normal distribution in training process, and we use the Adam optimizer with a learning rate of $1.0 \times 10^{-5}$.

\subsection{Local Fusion of Diffusion}
The average fusion method mentioned in Section~\ref{subsec:local_fusion} is defined as follows. 

\noindent \textbf{Average Fusion.} Suppose $x^k_t(\mathbf{p}) \sim \mathcal{N}(\mu^k_t(\mathbf{p}), \Sigma^k_t(\mathbf{p}))$, we have:

\begin{equation}
    \begin{aligned}
        x_t(\mathbf{p}) \sim \mathcal{N}(\frac{1}{|\mathcal{G}(\mathbf{p})|}\sum_{k\in \mathcal{G}(\mathbf{p})}\mu_t^k(\mathbf{p}), \frac{1}{|\mathcal{G}(\mathbf{p})|^2} \sum_{k\in \mathcal{G}(\mathbf{p})} \Sigma^k_t(\mathbf{p})).
    \end{aligned}
\end{equation}
The rapidly decreasing variance impacts generation diversity and quality. We, therefore, propose a stochastic TSDF fusion algorithm.

\subsection{User Study}
We conduct two user studies on meshes from generation and reconstruction refinement in Section~\ref{sec-exp: generation} and~\ref{sec-exp:refine}, which are slightly different.

\noindent \textbf{Generation.} We use same metric as Text2Room~\cite{hollein2023text2room}: Completeness and Perceptual. In every page of the survey, the users scores one scene from one method by 1-5 points on these 2 metrics. Then we take an average score on each method.

\noindent \textbf{Reconstruction Refinement.} We employ more metrics here, including details, completeness, plane quality, and edge quality. To save the time of the users, we use ranking rather than scoring for each scene.
The feedback score $S_i$ for the $i$-th scene is computed as 
\begin{equation}
    S_i = \frac{1}{d_i}\sum_{j=1}^{d_i} s(r_{i,j}),
\end{equation}
where $r_{i,j}\in {1, 2, 3, 4}$ represents the ranking given by the $j$-th user for the $i$-th scene. The function $s(r)=4-r$ converts the ranking into a score, with the $r$-th rank worth $4-r$ score. $d_i$ is the total number of valid feedbacks for the $i$-th scene.
By summing up the scores across all scenes, we obtain the total score 
\begin{equation}
    S = \sum_{i=1}^{N} S_i
\end{equation}

\section{More Results on Scene Generation}
\begin{figure*}[!ht]
    \centering
    \includegraphics[trim={8cm 3cm 10cm 3cm},clip,width=0.95\textwidth]{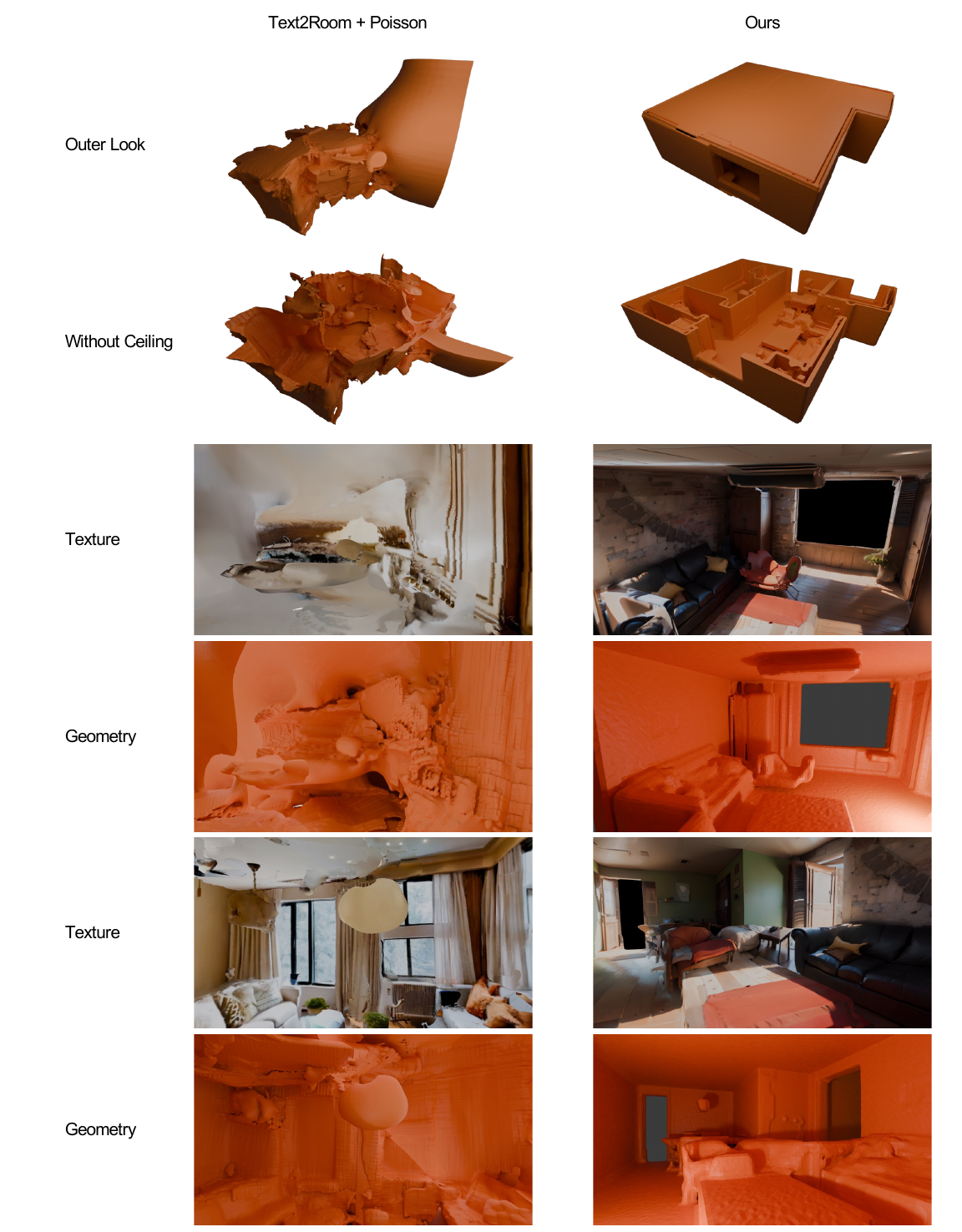}
    \caption{Comparison of Text2Room and our approach in larger views. As previous Fig.~\ref{fig:generated_scenes} shows, Poisson reconstruction significantly improves the performance of pure TextRoom, so that here we only demonstrate the results of Text2Room~\cite{hollein2023text2room} + Poisson~\cite{kazhdan2006poisson}. The textures of our results are produced by DreamSpace~\cite{yang2023dreamspace} as a post-processing of scene geometry generation.}
    \label{supfig:compare}
\end{figure*}

\begin{figure*}[!ht]
    \centering
    \includegraphics[trim={7cm 3.5cm 12cm 3.5cm},clip,width=0.97\textwidth]{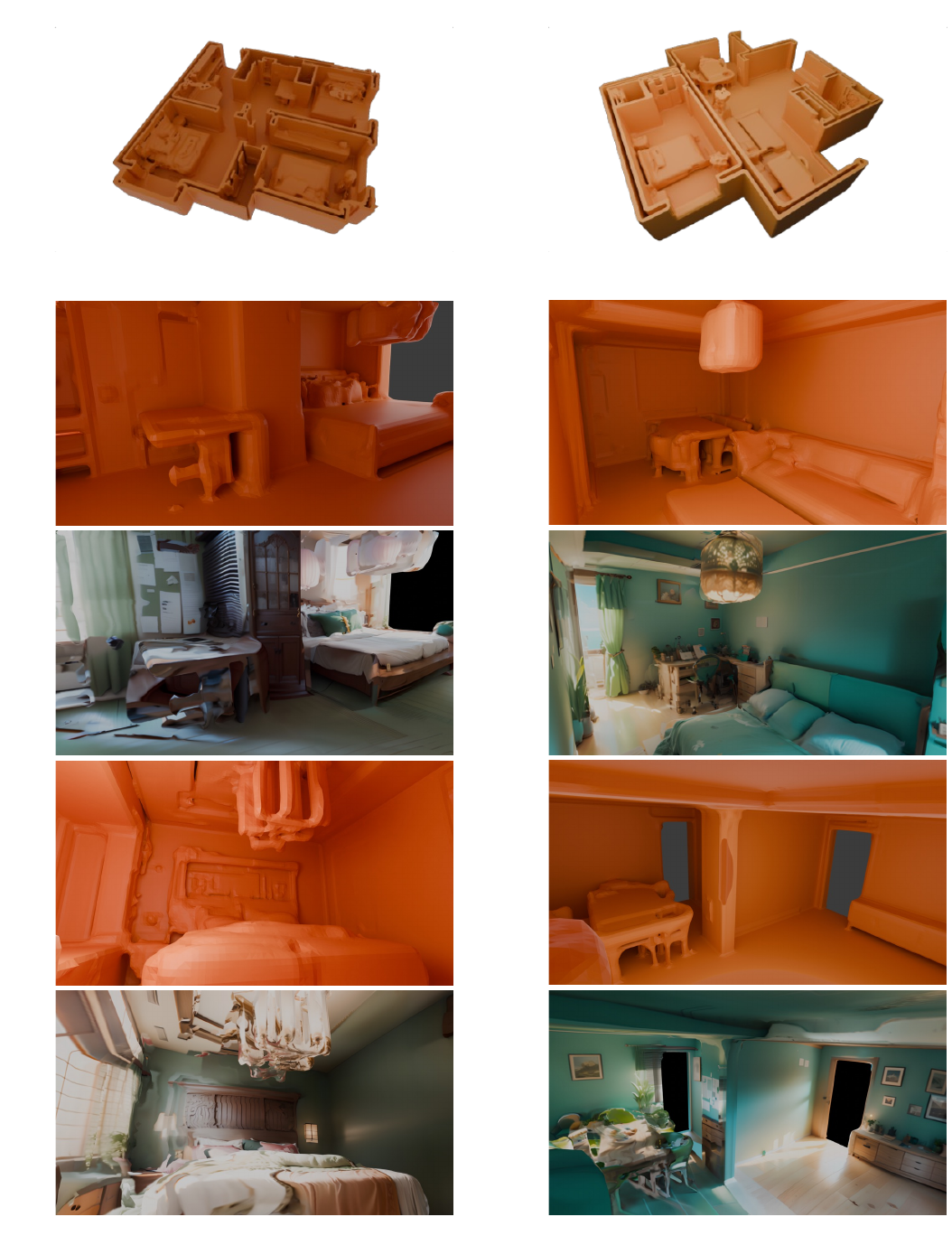}
    \caption{More generation samples in columns.}
    \label{supfig:more1}
\end{figure*}

\begin{figure*}[!ht]
    \centering
    \includegraphics[trim={7.5cm 4cm 12cm 3.5cm},clip,width=0.94\textwidth]{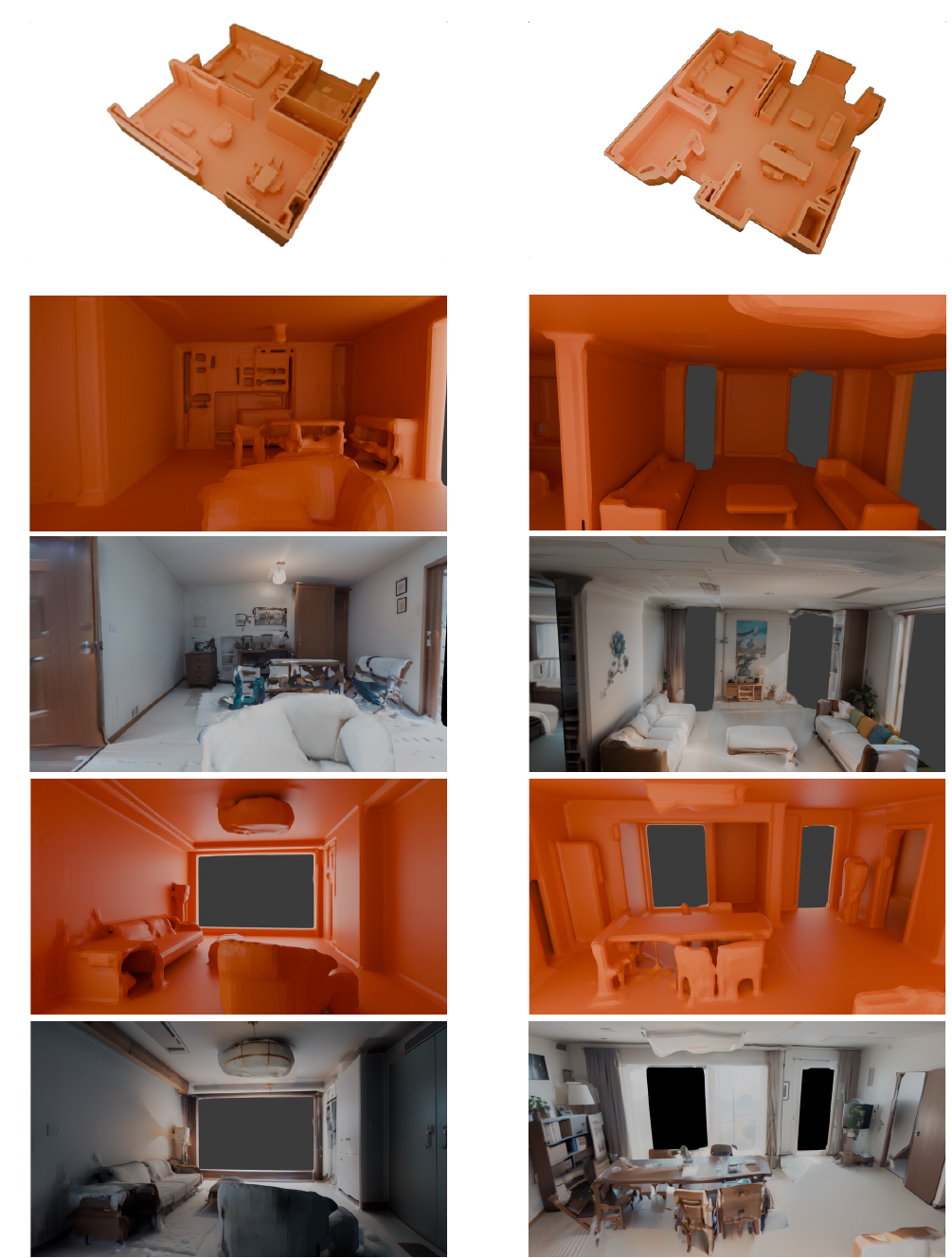}
    \caption{More generation samples in columns.}
    \label{supfig:more2}
\end{figure*}

We provide more scene generation samples as shown in Fig.~\ref{supfig:compare} - Fig.~\ref{supfig:more2}. 

Fig.~\ref{supfig:compare} is an additional comparison between our method and Text2Room~\cite{hollein2023text2room}. Since the Poisson~\cite{kazhdan2006poisson} reconstruction can produce better results than pure Text2Room, we only show the results of "Text2Room + Poisson". 
Fig.~\ref{supfig:more1} and Fig.~\ref{supfig:more2} are generated scene samples of our method.

